\documentclass[]{ceurart}

\usepackage{listings}
\usepackage{float}
\usepackage{placeins}
\usepackage{subcaption}
\usepackage{wrapfig}
\usepackage{caption}
\usepackage{pifont}
\usepackage{colortbl}
\usepackage[most]{tcolorbox}
\tcbuselibrary{most}
\usepackage{varwidth}

\definecolor{cyan10}{HTML}{E5F6FF}
\definecolor{cyan60}{HTML}{0072c3}
\definecolor{orange60}{HTML}{ba4e00}
\definecolor{yescolor}{RGB}{0, 150, 0}
\definecolor{nocolor}{RGB}{255, 0, 0}
\definecolor{mypink}{RGB}{255, 182, 193}
\definecolor{mypowderblue}{RGB}{176, 224, 230}
\definecolor{myframepink}{RGB}{219, 112, 147}
\definecolor{myframeblue}{RGB}{100, 149, 237}

\newcommand{\yes}{\textcolor{yescolor}{\ding{51}}}
\newcommand{\no}{\textcolor{nocolor}{\ding{55}}}

\begin{document}

\copyrightyear{2026}
\copyrightclause{Copyright for this paper by its authors.
  Use permitted under Creative Commons License Attribution 4.0
  International (CC BY 4.0).}

\conference{ASAIL 2026}

\title{Retrieval-Based Multi-Label Legal Annotation:\texorpdfstring{\\}{ }Extensible, Data-Efficient and Hallucination-Free}
\csgdef{casprelimstitle}{Retrieval-Based Multi-Label Legal Annotation: Extensible, Data-Efficient and Hallucination-Free}

\author[1]{Li Zhang}[%
orcid=0000-0002-7734-1644,
email=liz239@pitt.edu,
]
\author[2]{Jaromir Savelka}[%
orcid=0000-0002-8997-1799,
email=jsavelka@cs.cmu.edu,
]
\author[1]{Kevin Ashley}[%
orcid=0000-0003-1899-0973,
email=ashley@pitt.edu,
]
\address[1]{University of Pittsburgh, Pittsburgh, PA, USA}
\address[2]{Carnegie Mellon University, Pittsburgh, PA, USA}

\begin{abstract}
Multi-label legal annotation requires assigning multiple labels from large, evolving taxonomies to long, fact-intensive documents, often under limited supervision. Parametric encoders typically require task-specific training and retraining when the label set changes, while prompting generative large language models becomes costly and degrades as the label space grows. We cast legal annotation as retrieval: we embed documents and label descriptions with a frozen retrieval model and predict labels via \(k\)-nearest neighbors in the embedding space, enabling updates by re-embedding and re-indexing rather than gradient-based backpropagation. Across three legal datasets (ECtHR-A, ECtHR-B, and Eurlex with 100 labels), retrieval achieves competitive accuracy and strong data efficiency; on Eurlex, Qwen-8B retrieval improves Macro-F1 from 40.41 (GPT-5.2, zero-shot) to 49.12 while reducing estimated compute by \(\sim\)20--30\(\times\) compared to fine-tuning. With only \(N=100\) training samples, retrieval nearly doubles Micro-F1 over hierarchical Legal-BERT on ECtHR-A (48.29 vs.\ 27.87). We also quantify a reliability failure mode of generative inference: GPT-5.2 hallucinates labels outside the provided taxonomy in 0.12--0.9\% of test samples under deterministic decoding. In contrast, retrieval strictly respects defined label sets, eliminating hallucination by design. These results suggest retrieval-model-based annotators are a practical, deployable alternative for high-cardinality and rapidly changing legal label spaces.
\end{abstract}


\maketitle

\section{Introduction}

Real-world legal scenarios often present a dual dilemma: the need to navigate thousands of statutes, regulations, and doctrinal categories, and the challenge of working with scarce labeled data. Multi-label legal annotation sits at the intersection of these challenges, requiring systems that can (i) understand long, fact-intensive documents and (ii) operate under high-cardinality, long-tailed, and evolving label spaces. These requirements arise in practical settings such as case outcome prediction, charge/issue tagging, and topical assignment of legislation, where each document is associated with multiple legal concepts simultaneously \cite{ashley2017artificial,chalkidis2022lexglue,aletras2016predicting}.

\begin{figure}[ht]
  \centering
  \includegraphics[width=\linewidth]{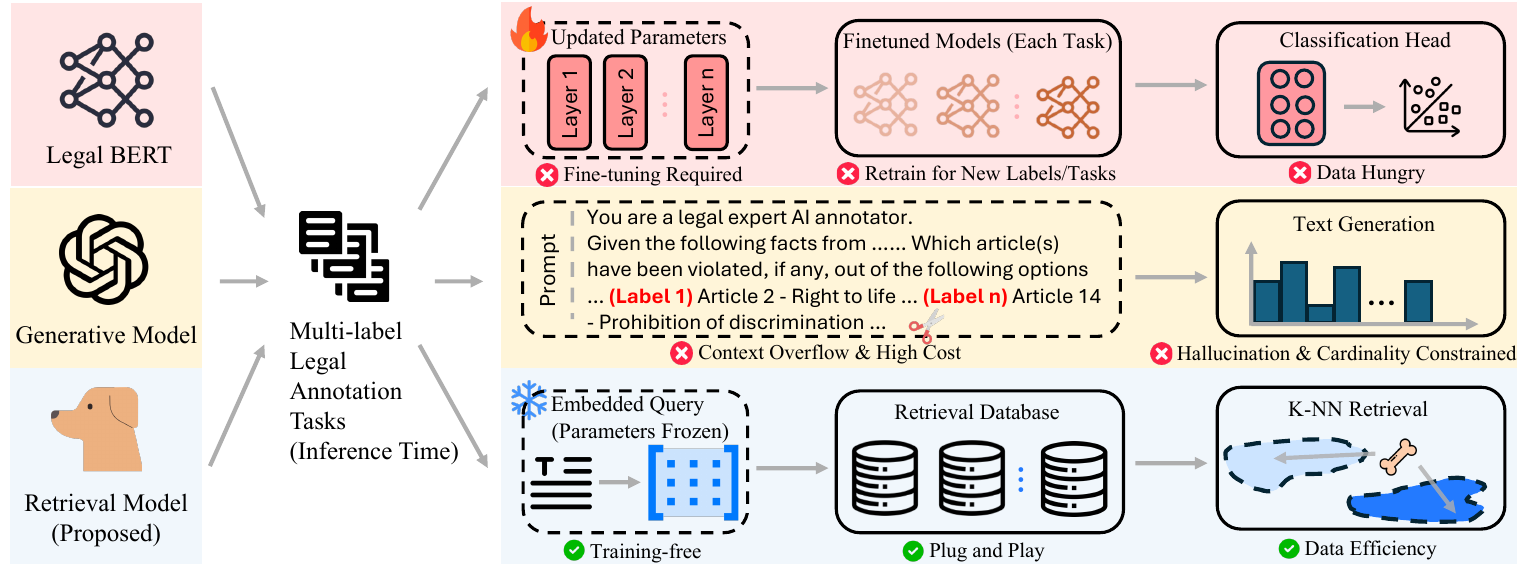}
  \caption{Comparison of Legal Annotation Paradigms (Inference Time). (a) \textbf{Parametric Fine-tuning (BERT)}: Requires updating model weights, data-hungry, rigid. (b) \textbf{Generative Zero-shot (GPT-5.2)}: Context window limited, expensive, slow. (c) \textbf{Proposed Retrieval Model (Qwen-3 Embedding)}: Retrieval-based, plug-and-play, handles a large and evolving set of labels efficiently.}
  \label{fig:comparison}
\end{figure}

Benchmarks highlight the structural difficulty of the problem. Observations from LexGLUE \cite{chalkidis2022lexglue} reveal that BERT-based methods consistently underperform on multi-label tasks compared to their single-label counterparts. Unlike single-label annotation, which forces a document into a mutually exclusive category (e.g., binary outcome prediction), multi-label annotation reflects the complexity of legal practice, where a single case or contract often implicates multiple statutes, regulatory domains, and doctrinal issues simultaneously.

However, datasets like EURLEX57K \cite{chalkidis2019large,chang2020taming} exacerbate this challenge with thousands of topic labels and severe imbalance. Moreover, legal taxonomies drift over time and across jurisdictions and languages \cite{chalkidis2021multieurlex}, making ``train once, deploy forever'' an unrealistic assumption. As a result, an ideal annotator should be accurate under limited supervision, stable under distribution shift, and cheap to adapt when the label set or support corpus changes.

Existing paradigms for legal annotation exhibit significant limitations. \textbf{Parametric models} based on pre-trained encoders (e.g., Legal-BERT) achieve strong performance after domain adaptation and task-specific fine-tuning \cite{chalkidis2020legal}. However, they are data-hungry and brittle in the long tail: rare labels and few-shot regimes lead to unstable decision boundaries, while adding new labels or updating the training distribution typically requires retraining.

\textbf{Generative LLMs} offer an alternative via in-context learning \cite{brown2020language} and prompting-based annotation \cite{savelka2023unreasonable,lee2025efficient}. Yet for multi-label classification with hundreds or thousands of labels and long documents, prompting can run into an intrinsic \emph{context bottleneck}: label definitions and calibration examples must fit within the prompt budget, and performance can degrade when relevant information is buried in long contexts \cite{liu2023lost,zhang2025llms}.

In this paper, we propose a solution based on \textbf{Retrieval Models}, specifically leveraging modern decoder-based embedding models and non-parametric inference. Our approach utilizes the semantic knowledge of large pre-trained models but projects it into a retrieval space, where annotation is performed by similarity search and \(k\)-Nearest Neighbors (k-NN) voting \cite{cover1967nearest}. This reframes classification as retrieval: new documents, new labels (with textual descriptions), and new evidence can be incorporated by re-embedding and re-indexing, rather than updating model weights.

Our contributions are threefold. First, we reframe legal multi-label classification as a scalable retrieval task. Second, we demonstrate that retrieval models can act as a \emph{universal annotator} that resolves the trade-off between cardinality and data availability, particularly beyond a ``cardinality threshold'' where prompting-based approaches degrade and may hallucinate labels. Third, we validate the superior efficiency and deployability of our approach, showing it is a viable on-premise alternative for sensitive legal data.

\FloatBarrier
\section{Related Work}

\subsection{Legal Text Annotation}
Legal text annotation underpins many AI \& Law applications, from charge and article prediction \cite{chi2025universal} to contract review \cite{hendrycks2021cuad}. In practice, legal annotation serves as a critical decision-support tool: it maps complex factual narratives to strictly controlled vocabularies---statutory provisions, treaty articles, or standardized taxonomies such as EuroVoc---enabling downstream applications including legal search, compliance monitoring, and judicial decision support. In this setting, the choice of annotation paradigm has concrete legal implications. First, \emph{hallucination-free guarantees} are essential: generating a plausible but non-existent statute constitutes a validity error that can mislead practitioners. Second, \emph{taxonomic rigor} is non-negotiable: collapsing fine-grained EuroVoc descriptors into broader categories disrupts regulatory monitoring workflows. Third, \emph{data sovereignty} matters for sensitive legal data that cannot be sent to third-party APIs due to privacy regulations (e.g., GDPR) or attorney-client privilege.

With the rise of pre-trained language models (PLMs), domain-adapted encoders such as Legal-BERT became strong defaults for legal classification \cite{chalkidis2020legal}, and benchmarks like LexGLUE and LegalBench standardized evaluation \cite{chalkidis2022lexglue,guha2023legalbench}. Multi-label legal annotation is particularly challenging due to label imbalance \cite{wais2025learning}, inter-label correlations, and dynamic taxonomies.

\subsection{Large Language Models for Legal Analysis}
LLMs have enabled prompting-based legal analysis via in-context learning \cite{brown2020language,gray2024using,luo2025automating}. Prompting techniques that externalize intermediate reasoning (e.g., chain-of-thought) can further improve reliability \cite{wei2022chain,yao2023tree}. However, multi-label classification at scale stresses core limitations of generation-centric inference: prompts must encode label definitions and constraints; as label spaces grow, prompt budgets saturate. Generation also introduces latency and cost that scales with input and output length.

Retrieval-augmented generation (RAG) mitigates some limitations by retrieving evidence before generation \cite{lewis2020retrieval,bareham2025curb}. Yet RAG systems still face the problem that the model must reason over retrieved passages within a fixed context window. In contrast, our approach treats annotation itself as retrieval.

\subsection{Text Representation and Retrieval-based Classification}
Our work builds upon the shift from generation to retrieval-centric text representation. Dense retrieval methods learn encoders that map queries and documents into a shared vector space \cite{yang2025qwen3}; DPR popularized contrastive training for semantic retrieval \cite{karpukhin2020dense}. Sentence-level representation learning further improved embedding quality with SBERT \cite{reimers2019sentence} and SimCSE \cite{gao2021simcse}. At the systems level, scalable approximate nearest neighbor search libraries such as FAISS \cite{johnson2019billion} make vector-space inference practical at large scale.

\section{Methodology}

We formally define the multi-label legal annotation task. Let $\mathcal{D} = \{(x_i, Y_i)\}_{i=1}^N$ be a dataset where $x_i$ represents a legal document and $Y_i \subseteq \mathcal{L}$ is the subset of relevant labels from the label space $\mathcal{L} = \{l_1, \dots, l_K\}$. Our goal is to learn a mapping $f: \mathcal{X} \rightarrow 2^\mathcal{L}$ that predicts the label subset $\hat{Y}$ for a new document $x$.

\subsection{Parametric Fine-tuning (BERT Baseline)}
In the parametric approach, we employ an encoder-only architecture (e.g., Legal-BERT) parameterized by $\theta$. The model parameters $\theta$ are updated to minimize the Binary Cross-Entropy (BCE) loss:
\begin{equation}
    \theta^* = \arg\min_\theta \sum_{(x, Y) \in \mathcal{D}_{train}} \mathcal{L}_{BCE}(f_\theta(x), Y)
\end{equation}
The document $x$ is encoded into a hidden representation $h_x = \text{Encoder}_{\theta^*}(x)$. For the Hierarchical Legal-BERT baseline, the document is split into overlapping segments, each encoded independently by the base BERT model, then aggregated by a 2-layer Transformer Encoder module to produce a document-level representation $h_x$. A classification head projects $h_x$ to label logits:
\begin{equation}
    P(l_k | x; \theta^*, W) = \sigma(W_k \cdot h_x + b_k)
\end{equation}
where $\sigma$ is the sigmoid function and $b_k$ is the bias term for label $l_k$.

\subsection{Generative Inference (GPT-5.2)}
For Generative LLMs, we treat classification as a text generation task. We construct a prompt $P(x, \mathcal{L})$ containing the document text, candidate labels, and instructions. The model generates a response $r = \mathcal{M}(P(x, \mathcal{L}))$, and a parsing function extracts predicted labels: $\hat{Y} = \phi(r)$.

\subsection{Proposed Retrieval-based Classification}
We propose a non-parametric retrieval framework. We utilize a decoder-only embedding model $E_\phi$ (e.g., Qwen-3 Embedding) to map both documents and label descriptions into a shared semantic vector space $\mathbb{R}^d$.

\begin{wrapfigure}{r}{0.5\textwidth}
  \centering
  \vspace{-12pt}
  \includegraphics[width=\linewidth]{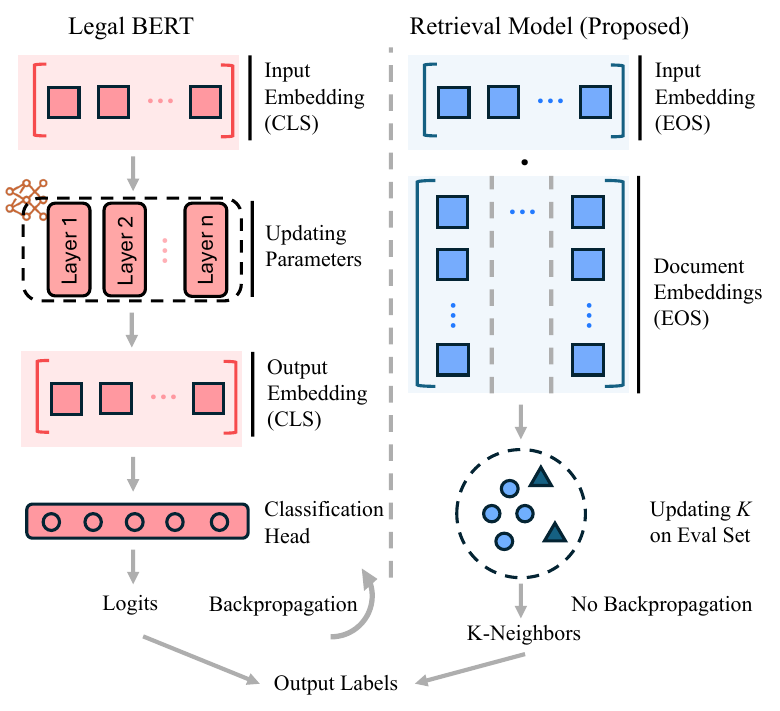}
  \caption{Training Phase Comparison. (Left) BERT updates all parameters $\theta$ via gradient descent. (Right) The Retrieval Model keeps the encoder frozen and selects $k$ on the validation set (no gradient updates).}
  \label{fig:training}
  \vspace{-10pt}
\end{wrapfigure}

\paragraph{Index Construction.}
We do not update the model weights $\phi$. Instead, we construct a semantic index $\mathcal{I}$ of label embeddings and select the hyperparameter $k$ on the validation set:
\begin{equation}
    k^* = \arg\max_k \text{F1}(\text{kNN}(k, \mathcal{D}_{val}, \mathcal{I}))
\end{equation}
Adding a new label simply involves computing its embedding and updating the index.

\paragraph{Inference.}
Let $d_{l_k}$ be the textual description of label $l_k$. The embeddings are:
\begin{equation}
    \mathbf{v}_x = E_\phi(x), \quad \mathbf{v}_{l_k} = E_\phi(d_{l_k})
\end{equation}
Cosine similarity determines the prediction:
\begin{equation}
    \hat{Y} = \text{Top-}k^*\left(\left\{\frac{\mathbf{v}_x \cdot \mathbf{v}_{l_k}}{\|\mathbf{v}_x\| \|\mathbf{v}_{l_k}\|}\right\}_{k=1}^K\right)
\end{equation}

\subsection{Theoretical Efficiency Analysis}
\label{sec:theoretical_efficiency}
Fine-tuning requires backpropagating through the full 8192-token context:
$C_{FT} \approx 6 \times N_{params} \times S_{samples} \times E_{epochs} \times L_{seq}$.
Retrieval involves only inference and indexing:
$C_{Ret} \approx 2 \times N_{params} \times S_{test} \times L_{seq}$.
Empirically, LoRA fine-tuning requires $\approx 3.8 \times 10^{16}$ FLOPs, whereas Retrieval inference requires only $\approx 1.9 \times 10^{15}$ FLOPs---a 20$\times$ efficiency improvement.

\section{Experiments}

We evaluate our method on datasets representing a spectrum of complexity: ECtHR-A (Judgment, 10 labels), ECtHR-B (Allegation, 10 labels), and Eurlex (Topic Induction, 100+ labels).

\begin{table}[h]
  \caption{Dataset statistics. $|L|$: label cardinality, $\bar{L}$: average labels per document.}
  \label{tab:datasets}
  \begin{tabular}{lccccc}
    \toprule
    Dataset & $N_{train}$ & $N_{test}$ & $|L|$ & $\bar{L}$ \\
    \midrule
    ECtHR A & 9,000 & 1,000 & 10 & 1.16 \\
    ECtHR B & 9,000 & 1,000 & 10 & 1.45 \\
    Eurlex & 55,000 & 5,000 & 100 & 4.51 \\
    \bottomrule
  \end{tabular}
\end{table}

\subsection{Implementation Details}
We extend Legal-BERT (\texttt{nlpaueb/legal-bert-base-uncased}, 110M parameters) to 8,192 tokens using a hierarchical architecture: documents are split into 64 segments of 128 tokens, encoded independently, and aggregated via a 2-layer \texttt{nn.TransformerEncoder} (\texttt{nhead}=8, \texttt{dim\_feedforward}=2048). The global aggregation layer is randomly initialized. The classification head uses \texttt{BCEWithLogitsLoss} with Micro-F1 as the model selection criterion. LoRA is applied only to the local Legal-BERT encoder (query and value projections, $r$=8, $\alpha$=16).

For Qwen Retrieval Models (0.6B, 4B, 8B), we chose Qwen-3 Embedding because it offers models at multiple scales sharing the same architecture, enabling controlled scaling analysis. The k-NN hyperparameter $k$ is tuned per dataset and model on the validation set, ranging from 5 to 20. GPT-5.2 is evaluated with deterministic decoding (\texttt{temperature=0}) and JSON-only output.

\section{Results}

\subsection{Efficiency}

\begin{table}[h]
  \caption{Computational efficiency analysis (FLOPs on 2,000 samples).}
  \label{tab:efficiency}
  \begin{tabular}{lccc}
    \toprule
    Method & FLOPs (Est.) & Rel. Cost & Mem. \\
    \midrule
    Legal-BERT (Full FT) & $5.7 \times 10^{16}$ & $\sim 30\times$ & $\sim 928$ MB \\
    Legal-BERT (LoRA) & $3.8 \times 10^{16}$ & $\sim 20\times$ & $\sim 50$ MB \\
    \textbf{Local Retrieval} & $\mathbf{1.9 \times 10^{15}}$ & $\mathbf{1\times}$ & \textbf{0 MB} \\
    \bottomrule
  \end{tabular}
\end{table}

\subsection{Accuracy vs.\ Label Cardinality}

\begin{table}[h]
  \caption{Main performance comparison (Micro-F1 / Macro-F1). Best results bolded.}
  \label{tab:results}
  \begin{tabular}{lcccccc}
    \toprule
    & \multicolumn{2}{c}{ECtHR-A} & \multicolumn{2}{c}{ECtHR-B} & \multicolumn{2}{c}{Eurlex} \\
    \cmidrule(lr){2-3} \cmidrule(lr){4-5} \cmidrule(lr){6-7}
    Model & Mi & Ma & Mi & Ma & Mi & Ma \\
    \midrule
    Legal-BERT & 76.25 & 65.62 & \textbf{81.98} & \textbf{78.38} & 60.03 & 24.40 \\
    GPT-5.2 & \textbf{79.33} & \textbf{76.31} & 73.95 & 73.00 & 43.05 & 40.41 \\
    \midrule
    Qwen-0.6B & 67.47 & 59.02 & 71.76 & 67.75 & 61.85 & 42.77 \\
    Qwen-4B & 70.44 & 61.05 & 74.63 & 69.94 & 62.97 & 44.71 \\
    Qwen-8B & 72.34 & 63.24 & 77.02 & 71.07 & \textbf{63.76} & \textbf{49.12} \\
    \bottomrule
  \end{tabular}
\end{table}

We observe a ``Cardinality Threshold'' phenomenon. \textbf{High Cardinality (Eurlex):} GPT-5.2's performance drops substantially (Macro-F1 40.41), consistent with difficulty attending to hundreds of label definitions simultaneously. Retrieval (Qwen-8B) achieves the best Macro-F1 (49.12). \textbf{Low Cardinality (ECtHR):} The best paradigm depends on task structure---fine-tuned encoders excel at pattern matching (ECtHR-B), generative models at deductive reasoning (ECtHR-A), while retrieval remains competitive without task-specific training.

\subsection{Data Scaling and Cold-Start Dynamics}

\begin{figure}[t]
  \centering
  \begin{subfigure}{\linewidth}
    \centering
    \includegraphics[width=\linewidth]{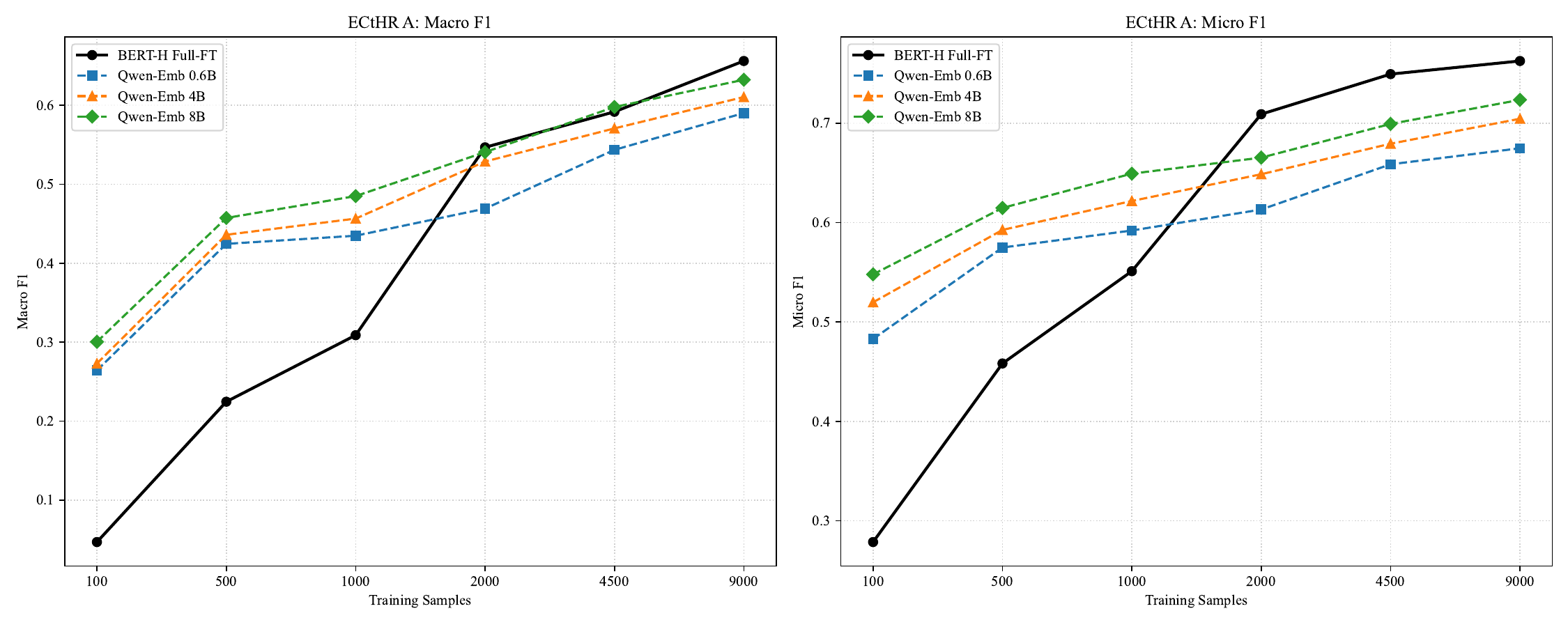}
    \caption{ECtHR-A Scaling Analysis}
  \end{subfigure}
  \vspace{0.5em}

  \begin{subfigure}{\linewidth}
    \centering
    \includegraphics[width=\linewidth]{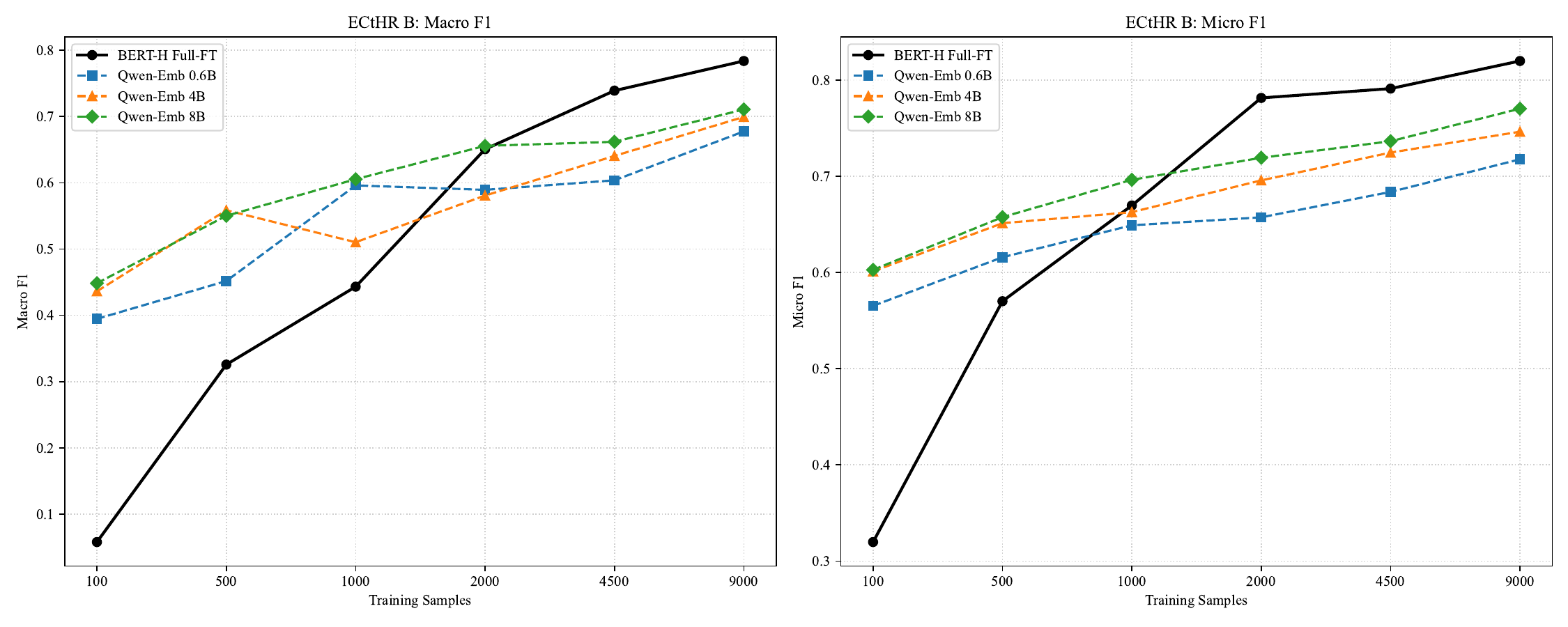}
    \caption{ECtHR-B Scaling Analysis}
  \end{subfigure}
  \vspace{0.5em}

  \begin{subfigure}{\linewidth}
    \centering
    \includegraphics[width=\linewidth]{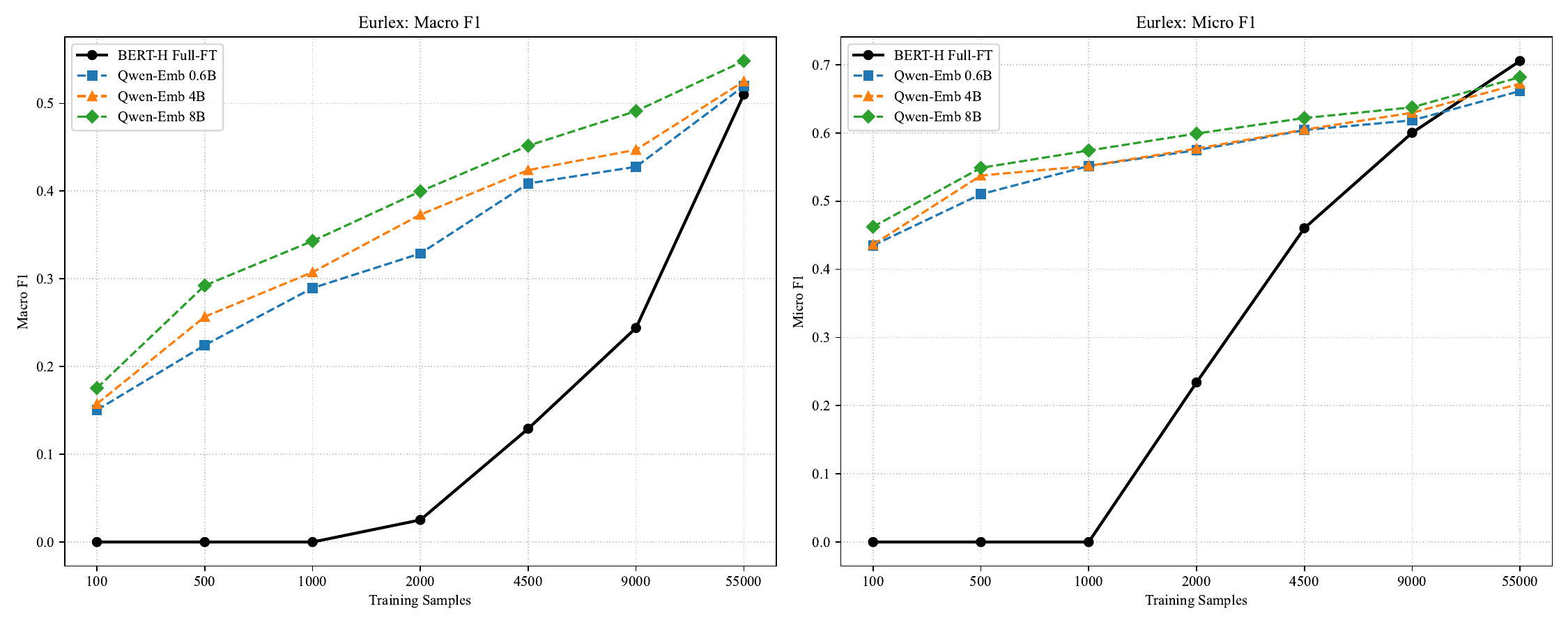}
    \caption{Eurlex Scaling Analysis}
  \end{subfigure}

  \caption{Performance scaling of Legal-BERT vs.\ Qwen Retrieval Models across varying training sample sizes.}
  \label{fig:scaling_analysis}
\end{figure}

Figure~\ref{fig:scaling_analysis} reports performance scaling across training-set sizes. \textbf{Cold-Start Win:} At $N=100$, Qwen-0.6B achieves nearly double the Micro-F1 of Legal-BERT on ECtHR-A (48.29 vs 27.87). \textbf{Convergence:} As data increases ($N > 2{,}000$), the gap narrows; Legal-BERT catches up on ECtHR tasks but retrieval maintains its lead on Eurlex even with full data. \textbf{Model Scaling:} Consistent improvement from 0.6B to 8B; even the smallest 0.6B model outperforms GPT-5.2 on Eurlex.

\section{Error Analysis}

\subsection{Quantifying Hallucinations}

We define a hallucination as any generated label $\hat{l} \notin \mathcal{L}_{candidate}$. Despite deterministic decoding and explicit constraints, GPT-5.2 produced hallucinations across all three datasets.

\begin{table}[h]
  \caption{Hallucination statistics across datasets.}
  \label{tab:hallucination_stats}
  \begin{tabular}{lccl}
    \toprule
    Dataset & Count & Rate & Top Hallucinated Labels \\
    \midrule
    ECtHR-A & 9 & 0.9\% & Article 7, Article 18 \\
    ECtHR-B & 7 & 0.7\% & Article 13, Article 7 \\
    Eurlex & 6 & 0.12\% & statistics, insurance \\
    \bottomrule
  \end{tabular}
\end{table}

In ECtHR tasks, GPT-5.2 generated valid ECHR articles excluded from the specific taxonomy---for instance, ``Article 7 -- No punishment without law.'' In Eurlex, hallucinations manifested as \emph{semantic drift}: broad topics such as ``statistics'' that do not exist in the granular EuroVoc subset. Table~\ref{tab:hallucination_example} illustrates a concrete case.

\begin{table}[h]
  \caption{Hallucination vs.\ Retrieval example. Input: \textit{``...collection of statistical data regarding external trade...''}}
  \label{tab:hallucination_example}
  \begin{tabular}{@{}p{0.12\textwidth}p{0.38\textwidth}p{0.38\textwidth}@{}}
    \toprule
    & \textbf{GPT-5.2 (Zero-shot)} & \textbf{Retrieval (k-NN)} \\
    \midrule
    Predicted & EU law \yes,~ external trade \yes,~ \textbf{statistics} \no & EU law \yes,~ external trade \yes,~ \textbf{economic analysis} \yes \\
    \midrule
    Analysis & ``statistics'' is semantically relevant but \emph{not in candidate list} (hallucination) & Nearest valid vector; ``statistics'' cannot be returned \\
    \bottomrule
  \end{tabular}
\end{table}

\subsection{Compliance and Safety of Retrieval}

Since k-NN selects from a fixed index $\mathcal{I}$, the probability of generating an invalid label is strictly zero ($P(\hat{l} \notin \mathcal{L}) = 0$). In practice, legal annotation systems operate as \emph{decision-support tools} within human-in-the-loop workflows. When a model proposes candidate labels, a legal professional reviews the output before it enters a case management system. With retrieval, the reviewer only needs to verify \emph{relevance}; with a generative model, the reviewer must also verify \emph{existence}---significantly increasing cognitive load and the risk of ``automation bias.''

\section{Discussion}

\subsection{The Practical Meaning of F1 Scores in Legal Practice}

Legal annotation systems are deployed as \emph{decision-support tools} for human experts. A Micro-F1 of 70 on Eurlex means the system narrows 100 potential descriptors to a relevant shortlist, transforming an hours-long manual search into a minutes-long verification task. Because retrieval guarantees zero hallucinations, residual errors are only omissions or sub-optimal matches, never fabricated labels requiring tedious fact-checking.

\subsection{From Prompt Engineering to Vector Space Geometry}

Our findings challenge the prevailing ``Generative-First'' orthodoxy in Legal NLP. Generative models treat classification as sequence generation, bounded by the context window; as the label space grows, encoding definitions into a prompt becomes a bottleneck \cite{liu2023lost}. Retrieval Models instead project legal concepts into a high-dimensional vector space where capacity is virtually unlimited and adding a new law is $O(1)$. The hallucination-free guarantee and data sovereignty are \emph{architectural} properties of retrieval that remain valuable regardless of how generative capabilities evolve. We view retrieval and generation as \emph{complementary} paradigms, and our Cardinality Threshold provides a guide for selecting the appropriate paradigm.

\subsection{Aligning AI with the Principle of Legality}

The \textit{Principle of Legality} mandates that decisions be based strictly on existing statutes. Our Error Analysis revealed that Generative LLMs, despite strict instructions, invent plausible but non-existent articles, violating the closed-world assumption of legal codification. A retrieval model may retrieve an \textit{incorrect} law (a relevance error) but will never \textit{fabricate} one (a validity error). When a lawyer reviews retrieval output, they only need to verify \textit{relevance}, reducing cognitive burden and the risk of ``automation bias.''

\subsection{Towards Modular and Sovereign Legal AI}

Our results suggest a shift from ``Monolithic'' to ``Modular'' Legal AI: a specialized 8B retrieval model outperforms a generalist trillion-parameter model while consuming $\sim$30$\times$ less compute \cite{belcak2025small}. This enables fully on-premise deployment for sensitive data under GDPR or client privilege, reserving heavy generative models for creative tasks (e.g., drafting arguments \cite{gray2025generating,zhang2025mitigating}) while classification is handled by efficient, sovereign retrieval systems.

\section{Conclusion}

We show that retrieval-based annotators balance efficiency, accuracy, and adaptability under evolving legal taxonomies, identifying a ``Cardinality Threshold'' where retrieval dominates over prompting. Retrieval offers strong cold-start performance and plug-and-play updates via re-embedding rather than retraining. GPT-5.2 hallucinates labels outside the taxonomy in 0.12--0.9\% of test samples; retrieval eliminates this failure mode by construction.

\section{Limitations and Future Work}

Our evidence is based on three English datasets ($K \le 100$); extending to multilingual settings \cite{chalkidis2021multieurlex} and additional legal systems \cite{goebel2025international,xiao2018cail2018} would strengthen external validity. The Cardinality Threshold is empirical: higher cardinality correlates with lower per-class sample counts, making it difficult to disentangle cardinality from data scarcity. Future work should explore supersampling to isolate these factors \cite{kaplan2020scaling,hoffmann2022empirical} and test transfer to other controlled-vocabulary domains such as medical coding \cite{mullenbach2018explainable} and contract classification \cite{tuggener2020ledgar}.

\section*{Reproducibility}
All experimental code and configurations are publicly available at \url{https://github.com/lizhang-AIandLaw/Retrieval-based-Multi-label-Legal-Annotation}.

\bibliography{reference}

\end{document}